%% file: main.tex
\input{src/setting}
\input{src/author}

\input{src/title}

\begin{document}
\maketitle
\input{content/0_abstract}
\input{content/1_introduction}
\input{content/2_related_work}
\input{content/3_preliminary}
\input{content/4_proposed_method}

\input{content/5_exp_setup}
\input{content/6_exp_result}
\input{content/7_discussion}

\input{content/8_conclusion}

\bibliography{src/library}
\bibliographystyle{ieeetr}
\end{document}

%% file: src/setting.tex
\documentclass[letterpaper, 10 pt, journal, twoside]{IEEEtran} 
\usepackage{caption}
\usepackage[utf8]{inputenc}
\usepackage[T1]{fontenc}

\usepackage{graphics} % for pdf, bitmapped graphics files
\usepackage{epsfig} % for postscript graphics files
\usepackage{bm}
\usepackage{times} % assumes new font selection scheme installed
\usepackage{amsmath} % assumes amsmath package installed
\usepackage{amssymb}  % assumes amsmath package installed
\usepackage{float}
\usepackage[space, compress]{cite}
\usepackage{mathtools}
\usepackage{graphicx}
\usepackage{hyperref}
\usepackage[font=footnotesize,labelfont=bf]{caption}
\usepackage{subcaption}
\usepackage{makecell}
\usepackage{upgreek}
\usepackage[dvipsnames]{xcolor}
\usepackage{pbox}
\usepackage{soul}

\usepackage{enumitem}
\usepackage[version=4]{mhchem}
\usepackage{multirow}
\usepackage{setspace, lipsum}
\usepackage{babel}
\usepackage{blindtext}
\usepackage{booktabs}
\usepackage{threeparttable}

\newcommand{\real}[1]{\mathbb{R}^{#1}}
\newcommand{\bs}[1]{\boldsymbol{#1}}
\newcommand{\prob}[1]{p\left(#1\right)}
\newcommand{\bst}[2]{\boldsymbol{#1}_{#2}}
\newcommand{\hst}[2]{\hat{\boldsymbol{#1}}_{#2}}
\newcommand{\dbst}[2]{\dot{\boldsymbol{#1}}{}_{#2}}
\newcommand{\sbrkt}[1]{\left(#1\right)}
\newcommand{\mbrkt}[1]{\left[#1\right]}
\newcommand{\lbrkt}[1]{\left\{#1\right\}}

\newcommand{\func}[3]{#1_{#2}\left(#3\right)}

\newcounter{qsubsection}
\newcommand{\qsubsection}[1]{%
  \refstepcounter{qsubsection}%
  \subsection*{Q\arabic{qsubsection}.\quad #1}%
}

%% file: src/author.tex
\author{Cheng-Yu Kuo$^{1}$, Hirofumi Shin$^{2}$, and Takamitsu Matsubara$^{1}$%
\thanks{$^{1}$Cheng-Yu Kuo and Takamitsu Matsubara are with Graduated School of Science and Technology, Nara Institute of Science and Technology, Nara 630-0192, Japan {\tt\footnotesize chengyu.kuo@naist.ac.jp; takam-m@is.naist.jp}}%
\thanks{$^{2} $Hirofumi Shin was with Honda R\&D, Ltd., Saitama 351-0114, Japan 
{\tt\footnotesize hirofumi.shin@ieee.org}
}%
}

%% file: src/title.tex
\title{Tracing Energy Flow:\\Learning Tactile-based Grasping Force Control \\to Reduce Slippage in Dynamic Object Interaction}

%% file: content/0_abstract.tex
\begin{abstract}
Regulating grasping force to reduce slippage during dynamic object interaction remains a fundamental challenge in robotic manipulation, especially when objects are manipulated by multiple rolling contacts, have unknown properties (such as mass or surface conditions), and when external sensing is unreliable. In contrast, humans can quickly regulate grasping force by touch, even without visual cues. Inspired by this ability, we aim to enable robotic hands to rapidly explore objects and learn tactile-driven grasping force control under motion and limited sensing. We propose a physics-informed energy abstraction that models the object as a virtual energy container. The inconsistency between the fingers’ applied power and the object’s retained energy provides a physically grounded signal for inferring slip-aware stability. Building on this abstraction, we employ model-based learning and planning to efficiently model energy dynamics from tactile sensing and perform real-time grasping force optimization. Experiments in both simulation and hardware demonstrate that our method can learn grasping force control from scratch within minutes, effectively reduce slippage, and extend grasp duration across diverse motion-object pairs, all without relying on external sensing or prior object knowledge. (Video: https://youtu.be/l3TJV29Mo6w)
\end{abstract}


%% file: content/1_introduction.tex
\section{Introduction}\label{sec:introduction}
\IEEEPARstart{G}{rasping} 
an object using fingertips to perform intended motion is common in tasks involving object manipulation or transport.
In these \emph{dynamic object interactions}, defined as scenarios where the object remains stationary relative to the finger frame while the fingers move, rolling contact occurs and slippage at the fingertips is often unavoidable~\cite{trinkle1990ijrr}. 
Under such conditions, object stability depends on adjusting fingertip forces to reduce slippage. However, especially in real-world scenarios, distinguishing slippage during dynamic object interaction is challenging due to the complex dynamics of multiple moving contacts, unknown object properties such as mass or surface condition, and unreliable external sensing caused by occluded vision.

\begin{figure}[!t]
    \centering
    \includegraphics[width=1\linewidth]{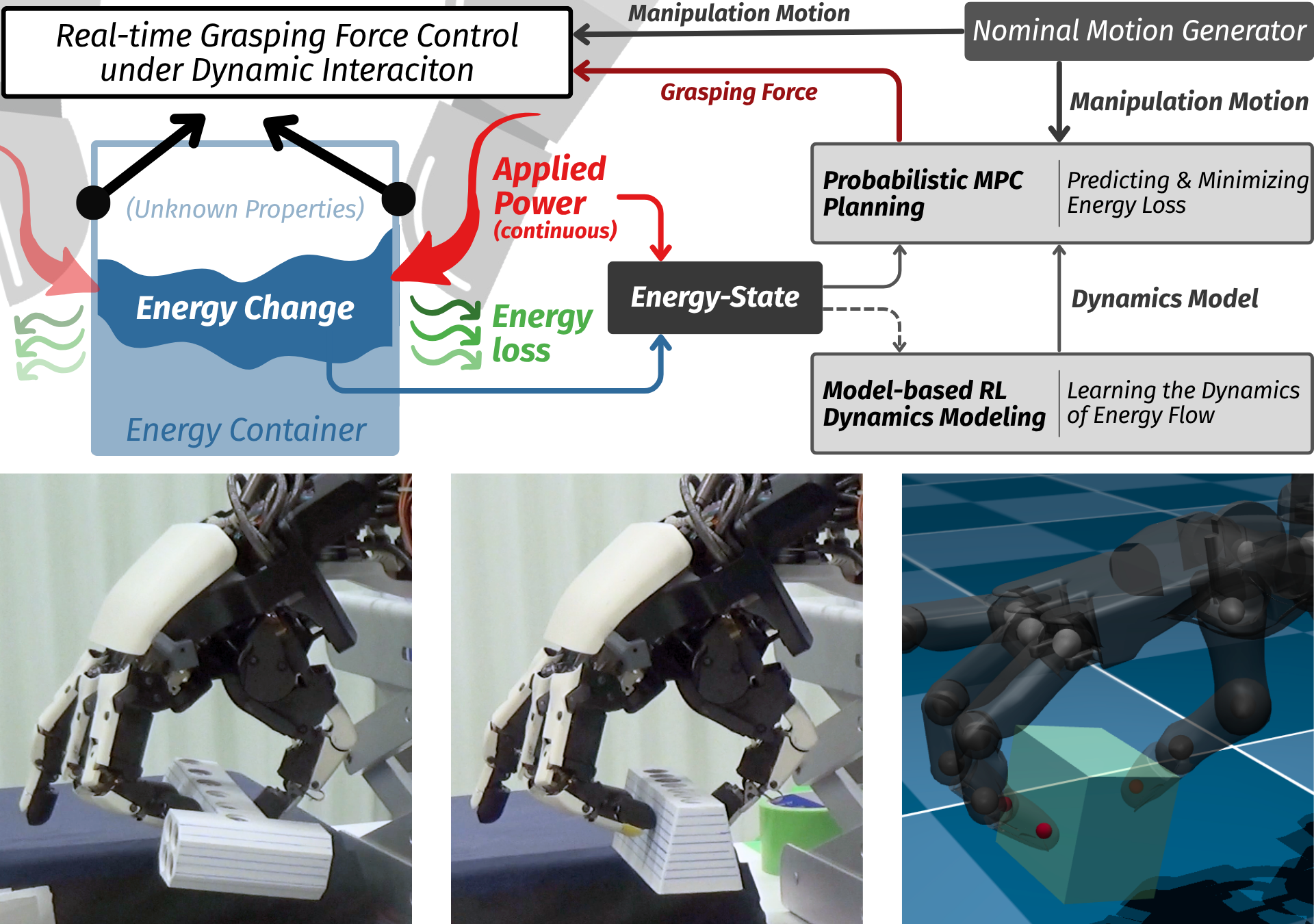}
    \caption{\textbf{Overview of the proposed framework for real-time grasping force control under dynamic object interaction} using tactile sensing. The object is abstracted as an energy container with unknown properties, where fingertip-applied power and retained energy change are compared to infer energy loss as a physically grounded indicator of slippage. These energy quantities are formulated into an energy-state representation, which is used in a Model-based Reinforcement Learning (MBRL) framework to learn energy-flow dynamics and perform grasping force optimization under dynamic object interaction via probabilistic Model Predictive Control (pMPC) to reduce slippage. The system is validated in both simulation and hardware.}
    \label{fig:intro}
\end{figure}
Despite these challenges during dynamic object interaction, human can rapidly regulate grasping force purely by touch. For example, when retrieving an unseen item from a bag, the sensation of slippage at the fingertips~\cite{schwarz2016tn} enables humans to adjust grasping force in real time. This highlights the unique role of touch and raises a central question: how can a robotic hand learn to adjust grasping force to minimize slippage during dynamic object interaction using only tactile sensing?

While principled solutions exist for real-time grasping force control~\cite{roa2014autonomous, bierbaum2008humanoids, ferrari1992icra}, they often fall short when facing real-world uncertainties such as unreliable external sensing and unclear object properties. Under these limitations, slippage must be inferred from tactile interaction alone. Furthermore, apart from tactile-based slippage sensing in static or quasi-static scenarios~\cite{komeno2024ral,abd2018aim,veiga2015iros}, motion and unknown surface conditions blur the boundary of slippage determination, making explicit supervision difficult~\cite{karl2018icra}. These challenges highlight the need for a tactile-driven approach that can reason about slippage without relying on explicit supervision.

To address these challenges, we propose a physics-informed energy abstraction that expresses slip-aware stability using only tactile sensing without requiring explicit supervision. Instead of modeling each fingertip’s contact dynamics, the object is abstracted as a virtual energy container. The system computes the total power applied by all fingertips and compares it to the change in the hypothetical object’s retained energy (due to unknown object properties). Under energy-conserving assumptions~\cite{folkertsma2017jtr}, consistency between the applied power and the change in retained energy indicates a stable interaction and yields a consistent estimate of the object's mass. Deviations from this consistency imply energy loss, which we interpret as potential slippage through physical reasoning. This abstraction provides a compact and physically grounded signal for slip-aware stability based solely on tactile sensing, making it well-suited for real-world scenarios with sensing uncertainty and unknown object properties.

To implement our energy abstraction for grasping force control, we integrate it into a Model-based Reinforcement Learning (MBRL) framework~\cite{deisenroth2011icml, kuo2023ral} for its sample efficiency and ability to support rapid on-site learning~\cite{valencia2023icra}, as illustrated in Fig.~\ref{fig:intro}. Specifically, we use our energy abstraction to construct an energy-state representation for MBRL to learn the dynamics of energy flow between finger-object interaction and perform real-time grasping force control using probabilistic Model Predictive Control (pMPC)~\cite{kamthe2018aistats, kuo2023ral}. In practice, we implement this approach within a modular system comprising a nominal motion generator, a grasping force planner, an impedance controller, and a model trainer. The nominal motion generator produces finger trajectories to induce object motion such as grasping, lifting, and manipulating. The grasping force planner adjusts fingertip forces online via pMPC to reduce slippage, while the impedance controller executes the combined motion and force commands. The model trainer updates the learned dynamics after each trial. 

We validate the system in both simulation~\cite{menagerie2022mujoco} and hardware~\cite{hasegawa2022icra}, demonstrating that our method can rapidly acquire grasping force control from scratch within minutes, effectively minimizing slippage across diverse motion and object conditions, without relying on external sensing or object knowledge. Our contributions are summarized as follows:
\begin{itemize}[leftmargin=*]
\item We propose a physics-informed energy abstraction that infers slippage from tactile-based energy inconsistency.
\item We integrate this abstraction with MBRL for real-time grasp force control.
\item We validate the approach in simulation and hardware without external sensing or prior object knowledge.
\end{itemize}

%% file: content/2_related_work.tex
\section{Related Work}
\subsection{Modalities in Dexterous Object Interaction}
Dexterous object interaction spans diverse research settings with varying sensing and support modalities. Some methods utilize external sensing (e.g., vision) to capture object states and guide control~\cite{morgan2022ral, morgan2019iros}, while others rely on geometric supports such as in-hand or palm-assisted stabilization~\cite{sundaralingam2018icra, song2018robio, melnik2021froroboticsandai, jiang2024contactmpc, funabashi2025focused}. Certain approaches combine both external sensing and geometric support to achieve complex tasks~\cite{openai2019rubiks, andrychowicz2019ijrr, chen2024dexforce, nagabandi2019corl, kumar2016icra}.

In contrast, we address dexterous interaction tasks without external sensing or geometric support, relying solely on fingertip tactile sensing to handle diverse objects and motions.

\subsection{Tactile Sensing and Slippage Reasoning}
Tactile sensing has been widely studied for slip detection using signal analysis~\cite{su2015humanoids}, supervised learning~\cite{veiga2015iros, abd2018aim}, or vibration-based methods~\cite{komeno2024ral}, which typically require explicit supervision. In dynamic manipulation, however, slippage emerges gradually, making segmentation and labeling difficult. Analytic approaches such as Damian et al.~\cite{damian2015asme} estimate 1D slip from artificial skin geometry and a force sensor, but are fundamentally limited in dynamic tasks where 3D slip occurs. Critically, the above methods are developed under quasi-static settings and have limited applicability to multi-finger dynamic object interactions, which involves object inertia and complex inter-finger forces, making slippage reasoning difficult.

Instead, we employ a physics-informed energy abstraction that integrates tactile sensing into model-based control. By estimating energy consistency from forces and velocities between multiple fingers and the object, our approach infers slippage based on physical laws, enabling label-free, real-time detection that accounts for object inertia and inter-finger reaction forces during dynamic manipulation.

\subsection{Energy-based Reasoning in Dexterous Manipulation}
Energy-based reasoning has been applied to dexterous manipulation, particularly with underactuated hands. Morgan et al.~\cite{morgan2019iros, morgan2022ral} introduce hand-centric energy formulations based on the mechanical compliance of an underactuated hand, using them to guide finger gaiting and grasp transitions.

Alternatively, we use object-centric energy abstraction that models the object as a virtual energy container and assesses grasp stability via energy consistency. This abstraction aggregates multi-finger tactile input into a compact signal, enabling slip-aware force control in real time.

\subsection{Model-based Learning in Dexterous Object Interaction}
Like our work, many recent studies explore the advantages of model-based learning in dexterous object interaction, particularly for managing complex dynamics and interaction forces in high-dimensional control tasks. For instance, Kumar et al.\cite{kumar2016icra} employ local linear models to achieve sample-efficient learning with a high-DOF robotic hand. Nagabandi et al.\cite{nagabandi2019corl} use deep ensemble dynamics with model predictive control to manipulate free-floating objects. MPC-SAC~\cite{omer2021iceee} blends offline policy learning with online planning for robustness, while Radosavovic et al.~\cite{radosavovic2021iros} propose state-only imitation learning to simplify model-assisted training.

While recent works demonstrate the efficiency of model-based learning in handling complex dynamics, we build on this advantage by introducing an energy abstraction that allows capturing tactile-driven and slip-aware interactions without modeling contact mechanics. This enables learning real-time grasping force control without relying on external sensing.

%% file: content/3_preliminary.tex
\section{Preliminaries}\label{sec:preliminary}
MBRL with probabilistic dynamics, such as Gaussian Processes, offers strong sample efficiency but suffers from limited real-time applicability due to high computational cost~\cite{deisenroth2011icml}. Therefore, we adopt the Fourier-featured Linear Gaussian Model (LGM-FF)~\cite{kuo2023ral}, which achieves scalable learning and high-frequency control while preserving the benefits of probabilistic dynamics modeling.

First, we view the system as a discrete-time transition: $\bst{x}{t+1} = f(\bst{x}{t}, \bst{u}{t}) + \epsilon$,
where $\bst{x}{t} \in \mathbb{R}^D$ is the system state, $\bst{u}{t} \in \mathbb{R}^U$ is the control input, and $\epsilon$ denotes Gaussian noise. 

Next, LGM-FF models each state dimension independently using input-target tuples $(\bst{x}{t}, \bst{u}{t}, \bst{y}{i})$, where $\bst{y}{i}$ is the $i$-th component of the next state $\bst{x}{t+1}$. This yields a probabilistic prediction $\prob{x_{i, t+1}} \approx \mathcal{N}(\mu_{i, t+1}, \Sigma_{i, t+1})$ for each dimension.

For multi-step planning, the predictive distribution at the next timestep is obtained via uncertainty propagation~\cite{kamthe2018aistats}: 
\begin{equation}
    \prob{\bst{x}{t+1}} = \mathbb{E}_{\prob{\bst{x}{t}}}[f(\bst{x}{t}, \bst{u}{t})] = \func{f}{M}{\prob{\bst{x}{t}}, \bst{u}{t}}. \label{eq:lgmff_moment_matching}
\end{equation}

With LGM-FF, the complexity of uncertainty propagation scales with $\mathcal{O}(DM^2)$, where $M$ is the number of random Fourier features, independent of the sample size $N$. Since $M \ll N$ in practice~\cite{kuo2023ral}, this significantly reduces planning time, enabling efficient real-time control.

%% file: content/4_proposed_method.tex
\begin{figure}[!t]
    \centering
    \includegraphics[width=1\linewidth]{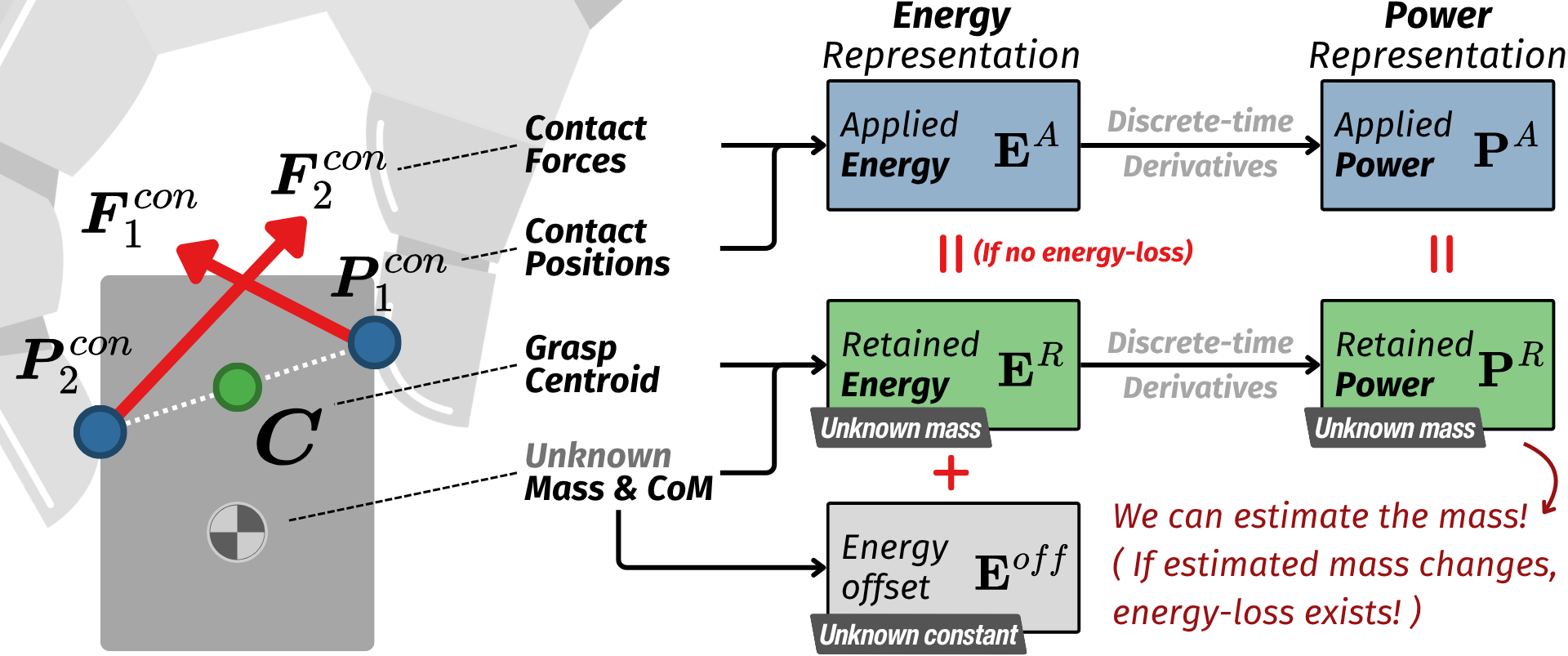}
    \caption{\textbf{Energy abstraction from multiple contact interaction.} We compute the total applied power from all fingertip contacts and compare it with the object’s retained power. Inconsistent mass estimates reflect energy loss, which can be attributed to slippage. This physics-informed abstraction enables slip-aware stability assessment without requiring object observation.}
    \label{fig:energy_power_overview}
\end{figure}

\section{Proposed Method}\label{sec:proposed_method}
We propose a physics-informed abstraction for object manipulation. By characterizing tactile-only finger-object interaction as energy flow, our method provides a compact and physically grounded signal to reason about slippage. This signal forms an energy-state used in MBRL to capture energy consistency and enable slip-aware control without requiring prior knowledge of object properties or external sensing. 
\subsection{Energy Abstraction for Finger-Object Interaction}
Here, we assume a rigid, fixed-shape object with unknown but constant mass $m$, and adopt point contacts (the instantaneous position of rolling contacts) for all energy computation.
To reason about contact interactions, we model the object as a virtual energy container, where the total energy $\mathbf{E}_t^A$ applied by $n$ fingers is estimated by integrating contact forces ($\bst{F}{}^{con}\in\real{3}$) and contact point velocities ($\dbst{P}{}^{con}\in\real{3}$):
\begin{equation}
    \mathbf{E}_t^A = \int_0^t \sum_{i=1}^{n} \bst{F}{i,t}^{con} \circ \dbst{P}{i,t}^{con} \, dt \in\real{3}. \label{eq:energy_applied}
\end{equation}
Also, we calculate the object's hypothetical retained energy $\mathbf{E}_t^R$ using the grasp centroid $\bst{C}{t}$ (mean of contact positions):
\begin{equation}
    \mathbf{E}_t^R = m \cdot [\bs{g}^\top\bst{C}{t},\ \frac{1}{2} \dbst{C}{t}^{\circ 2}]\in\real{4}. \label{eq:energy_retained}
\end{equation}
Under ideal energy conservation, the total applied and retained energies should differ by a constant offset:
\begin{equation}
     \bs{1}^\top \mathbf{E}_t^A - \bs{1}^\top \mathbf{E}_t^R = E^{off} = \text{const.}, \label{eq:energy_discrepancy}
\end{equation}
where $E^{\text{off}}$ accounts for the unknown energy offset between the object’s unknown center of mass and the grasp centroid. 

To eliminate this unknown offset, we take discrete-time derivatives of the applied and retained energies, and calculate the applied $\mathbf{P}_{t}^{A}$ and retained $\mathbf{P}_{t}^{R}$ power at each step as:
\begin{align}
    \mathbf{P}_{t}^{A} &= \mathbf{E}_{t}^{A} - \mathbf{E}_{t-1}^{A} \in\real{3},  \label{eq:power_applied} \\
    \mathbf{P}_{t}^{R} &= \mathbf{E}_{t}^{R} - \mathbf{E}_{t-1}^{R} = m \tilde{\mathbf{P}}_{t}^{R}\in\real{4}, 
    \label{eq:power_retained}
\end{align}
where $\tilde{\mathbf{P}}_{t}^{R}\in\real{4}$ denotes the ``massless'' retained power. 

Ideally, there is no discrepancy between the total applied and retained power; thus, the object’s mass is estimated as
\begin{equation}
    \tilde{m}_t = \bs{1}^\top \mathbf{P}_t^A \big/ \bs{1}^\top \tilde{\mathbf{P}}_t^R. \label{eq:mass_estimation}
\end{equation}
Under the constant mass assumption, $\tilde{m}_t$ should remain steady. In practice, slippage or uncoordinated forces cause energy loss, leading to fluctuations in the estimated mass. We interpret these fluctuations as physically grounded signals of slip-aware stability, as illustrated in Fig.~\ref{fig:energy_power_overview}.

\subsection{Dynamics Learning and Control with Energy Abstraction}
Utilizing the energy abstraction, we define an energy-state for modeling finger-object energy flow using MBRL:
\begin{equation}
    \bst{x}{t} := [\mathbf{P}_t^A,\ \tilde{\mathbf{P}}_t^R,\ \bst{\Theta}{t}] \in \mathbb{R}^{10},
\end{equation}
where $\bst{\Theta}{t}\in\real{3}$ is the orientation of the grasp centroid. 
The control signal $\bst{u}{t} \in \mathbb{R}^7$ comprises desired centroid velocities (linear $\dbst{C}{t}^\star\in\real{3}$ and rotational $\dbst{\Theta}{t}^\star\in\real{3}$) for motions and corresponding grasping force $F_t^\star\in\real{}$:
\begin{equation}
    \bst{u}{t} := [\dbst{C}{t}^\star, \dbst{\Theta}{t}^\star, F_{t}^\star]{}^\top \in\real{7}
\end{equation}

By learning the energy-state transition dynamics using LGM-FF, the grasping force is determined by a probabilistic Model Predictive Control (pMPC)~\cite{kuo2023ral}. At each step, a nominal motion generator produces reference velocity and angular velocity commands $(\dbst{C}{t}^\star,\ \dbst{\Theta}{t}^\star)$, while pMPC computes the grasping force sequence ($\bst{F}{t}^\star\in\real{H}$) that minimizes expected power discrepancy over a prediction horizon $H$:
\begin{equation}
\label{eq:pmpc_compact}
\min_{\bst{F}{t}^\star} \sum_{k=2}^{H+1} \mathbb{E}[\ell(\hst{x}{k})], \ \text{s.t. } \hst{x}{k+1} = \func{f}{M}{\hst{x}{k}, \hst{u}{k}},\ \hst{x}{1}=\bst{x}{t}
\end{equation}
where $\ell(\hst{x}{k})$ penalizes fluctuations in estimated mass as a proxy for energy inconsistency. The first entry in the optimized sequence is used as the applied force ($F_t^\star := \bst{F}{t}^\star(1)$). This force is then distributed among the contacting fingers via force closure~\cite{ferrari1992icra}, while ignoring object mass.

\subsection{Online Model-based Learning and Control}

We adopt an MBRL strategy that learns dynamics online through iterative manipulation trials. At each step, a nominal motion generator induces object motion, while pMPC optimizes the grasping force by minimizing expected energy loss based on predicted energy-state transitions. To mitigate control latency, we employ an ahead-planning scheme~\cite{kuo2023ral} that pre-computes the grasping force for the next step. Initial trials rely on random exploration, while subsequent executions leverage model-based control. After each trial, the dynamics model is updated using the collected samples, progressively improving prediction and control performance. This pipeline enables learning real-time grasping force control without requiring external sensing or knowing object properties.

%% file: content/5_exp_setup.tex
\begin{figure}[!t]
    \centering
    \includegraphics[width=1\linewidth]{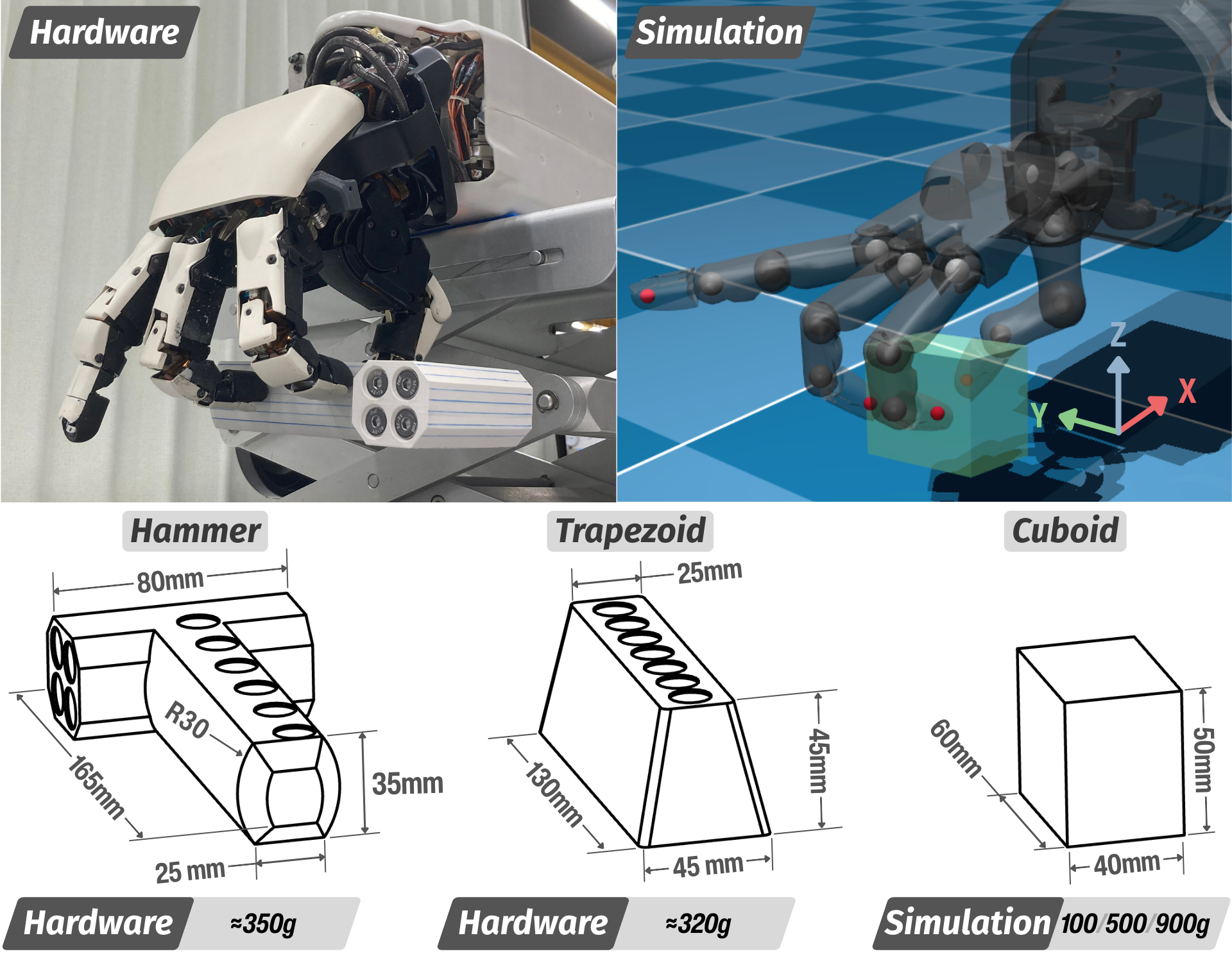}
    \caption{\textbf{Robotic hands and manipulated objects in the hardware and simulated environments.}}
    \label{fig:robot_setup}
\end{figure}
\begin{figure}[!t]
    \centering
    \includegraphics[width=1\linewidth]{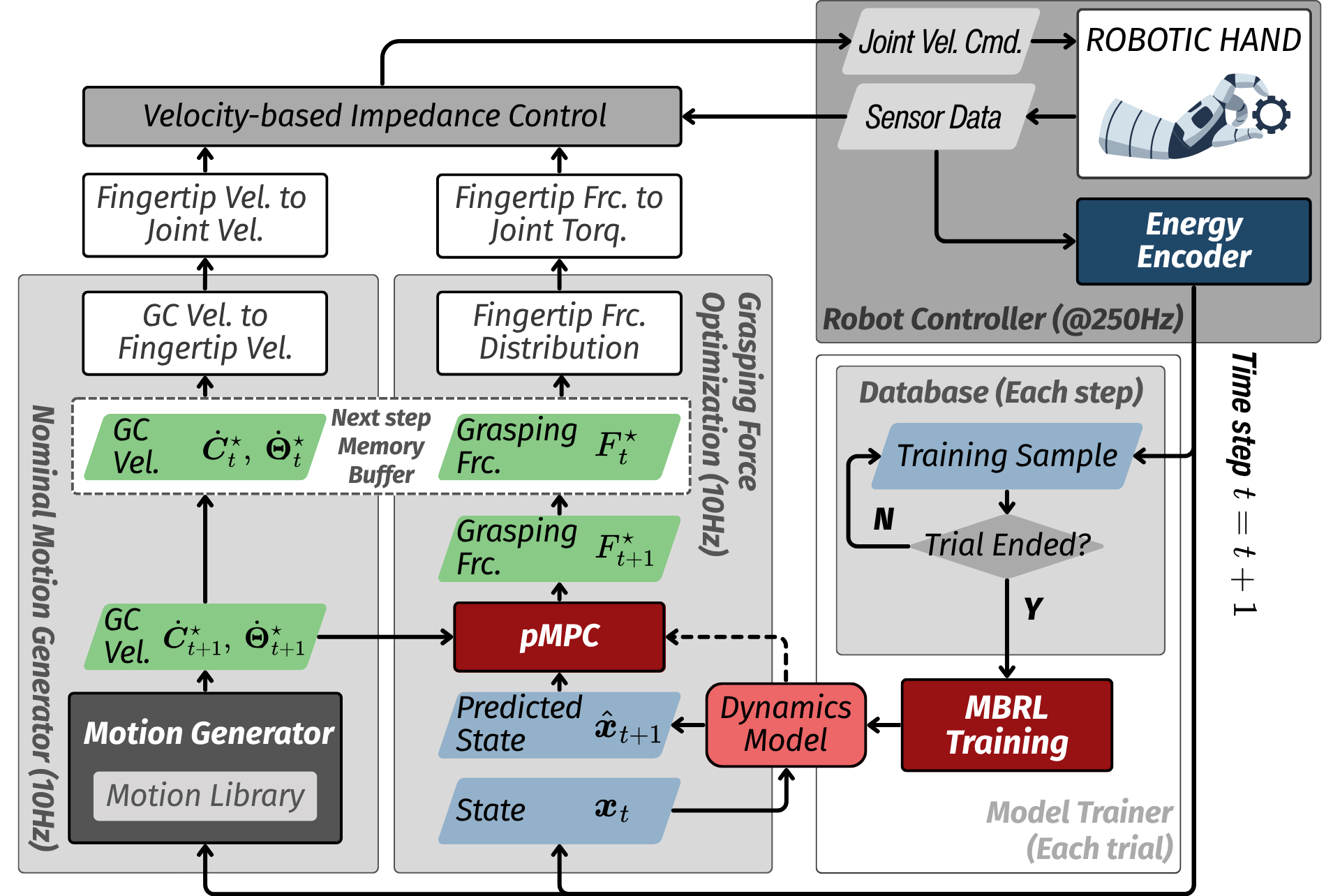}
    \caption{\textbf{Overview of our control framework integrating the energy abstraction and MBRL.} The system consists of a nominal motion generator, a grasping force planner, an impedance controller, and a model trainer. The motion generator produces baseline Grasp Centroid (GC) and finger trajectories to induce object motion. The energy abstraction constructs a compact energy-state representation from tactile sensing, which is used by the MBRL to learn interaction dynamics. Real-time grasping force optimization is performed by pMPC based on the learned energy dynamics. This structure supports real-time grasping force control with tactile sensing only.}
    \label{fig:control_framework}
\end{figure}
\begin{figure}[!t]
    \centering
    \includegraphics[width=1\linewidth]{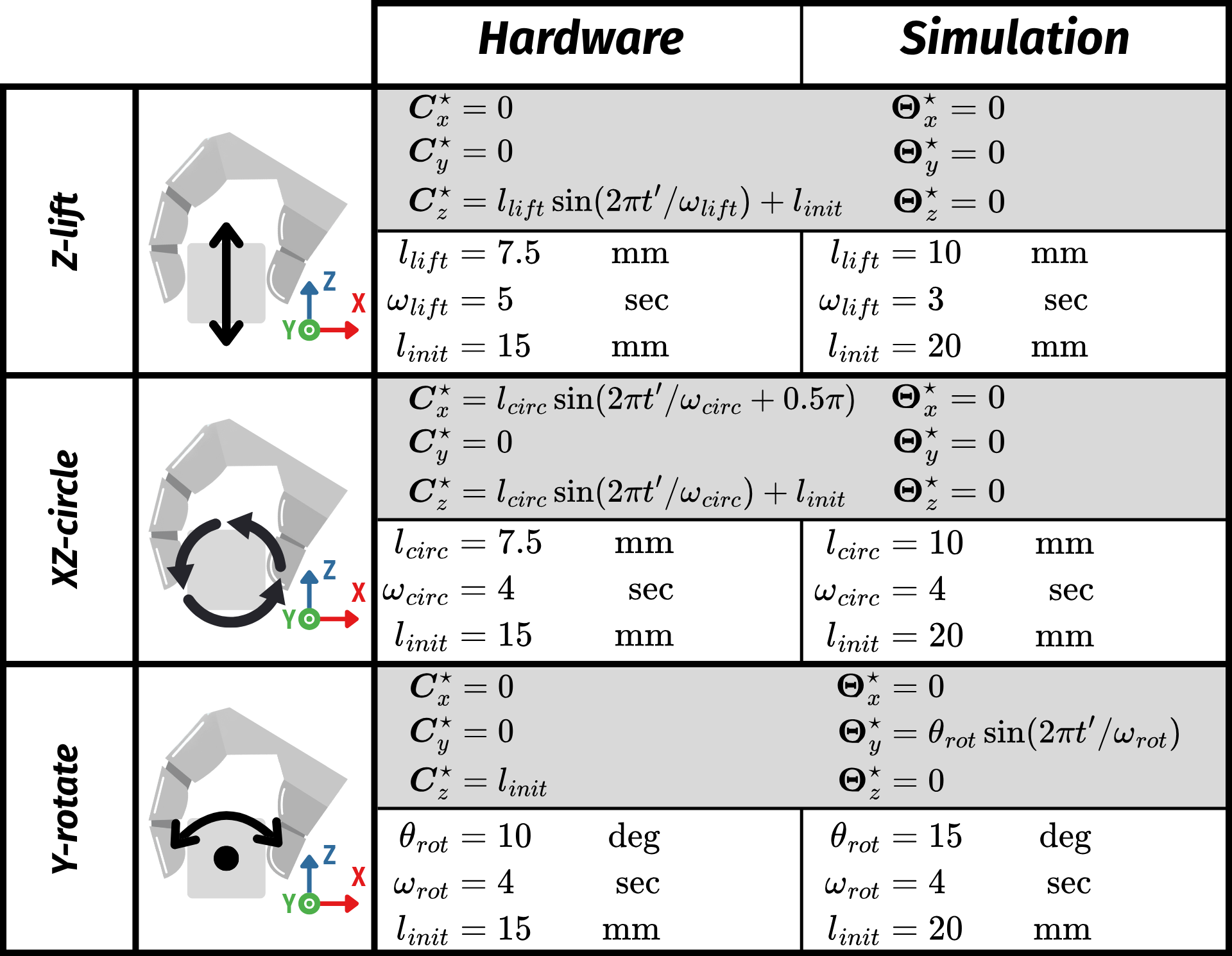}
    \caption{\textbf{Definition of periodic manipulation motions.} 
    The reference positions of the grasp centroid are denoted by $\boldsymbol{C}^\star$ (translation) and $\boldsymbol{\Theta}^\star$ (rotation). Each motion is described by a sinusoidal profile with angular frequency $\omega$ and amplitudes $l_{lift}$, $l_{circ}$, and $\theta_{rot}$ for Z-lift, XZ-circle, and Y-rotate respectively. The elevation offset $l_{\text{init}}$ denotes the lifted position during transition, and $t’ = t_a - 4$ indicates the relative time from the lift phase to the onset of manipulation. Parameters differ slightly between simulation and hardware to accommodate the kinematic constraints of each robot hand.}
    \label{fig:predefined_motion}
\end{figure}

\begin{figure}[!t]
    \centering
    \includegraphics[width=1\linewidth]{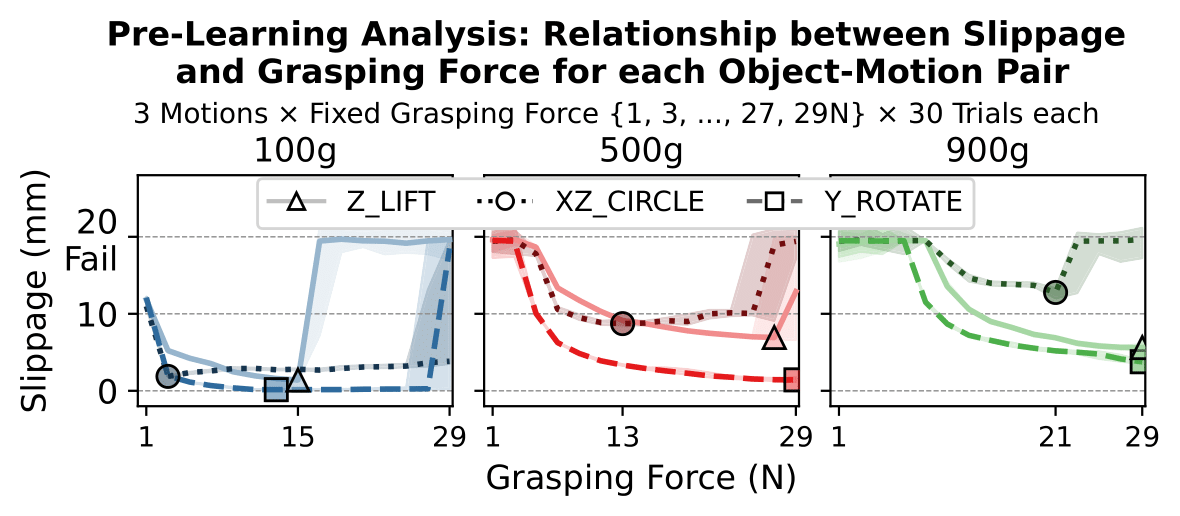}
    \caption{\textbf{Pre-learning analysis of slippage versus grasping force across object-motion pairs.} Each column shows slippage (the increased distance between finger and object frame) under three manipulation motions for objects of different masses, with fixed grasping forces from 1N to 29N over 30 trials. Solid lines show median slippage; shaded areas show range. Markers indicate the force minimizing slippage for each motion.}
    \label{fig:sim_slippage_force_compare}
\end{figure}

\begin{figure}[!t]
    \centering
    \includegraphics[width=1\linewidth]{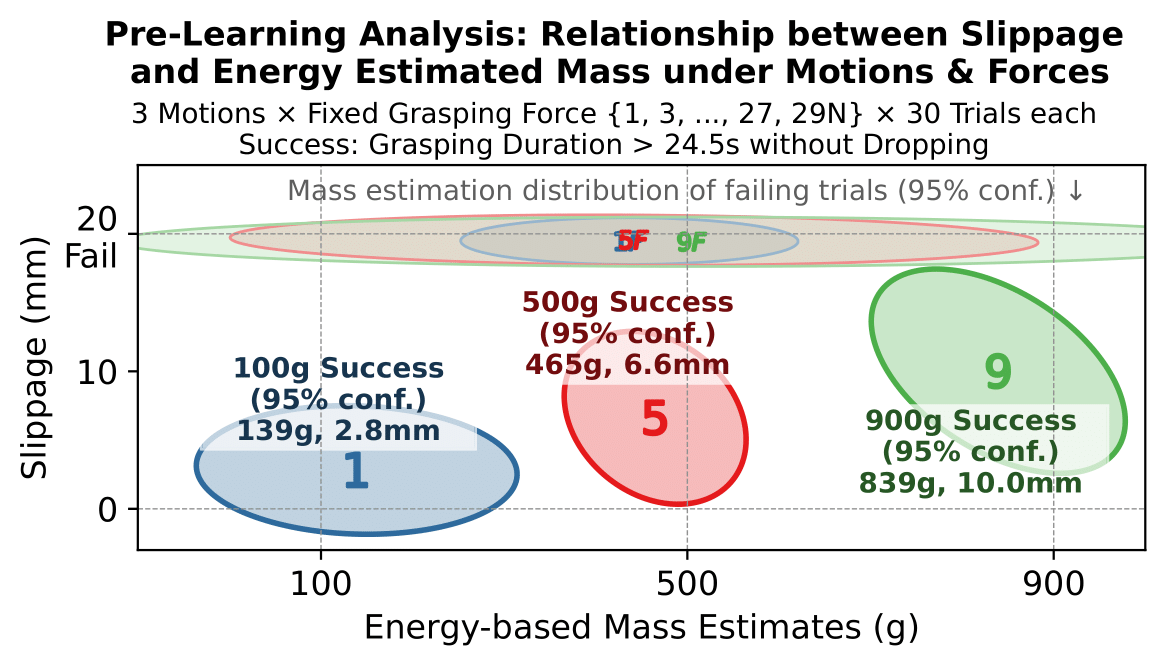}
    \caption{\textbf{Pre-learning analysis of energy-based mass estimation versus slippage.} Each ellipse shows the 95\% confidence ellipse of energy-based mass estimates and slippage (the increased distance between finger and object frame) for 100g, 500g, and 900g objects across \{15 fixed grasping forces $\times$ 3 motions $\times$ 30 trials\}. Trials are categorized as successful or failing. Successful trials (solid ellipses) produce mass estimates that are closer to ground truth with lower slippage, while failing trials (transparent ellipses) show larger estimation errors and higher slippage.}
    \label{fig:sim_slippage_mass_compare}
\end{figure}

\begin{figure}[!t]
    \centering
    \includegraphics[width=1\linewidth]{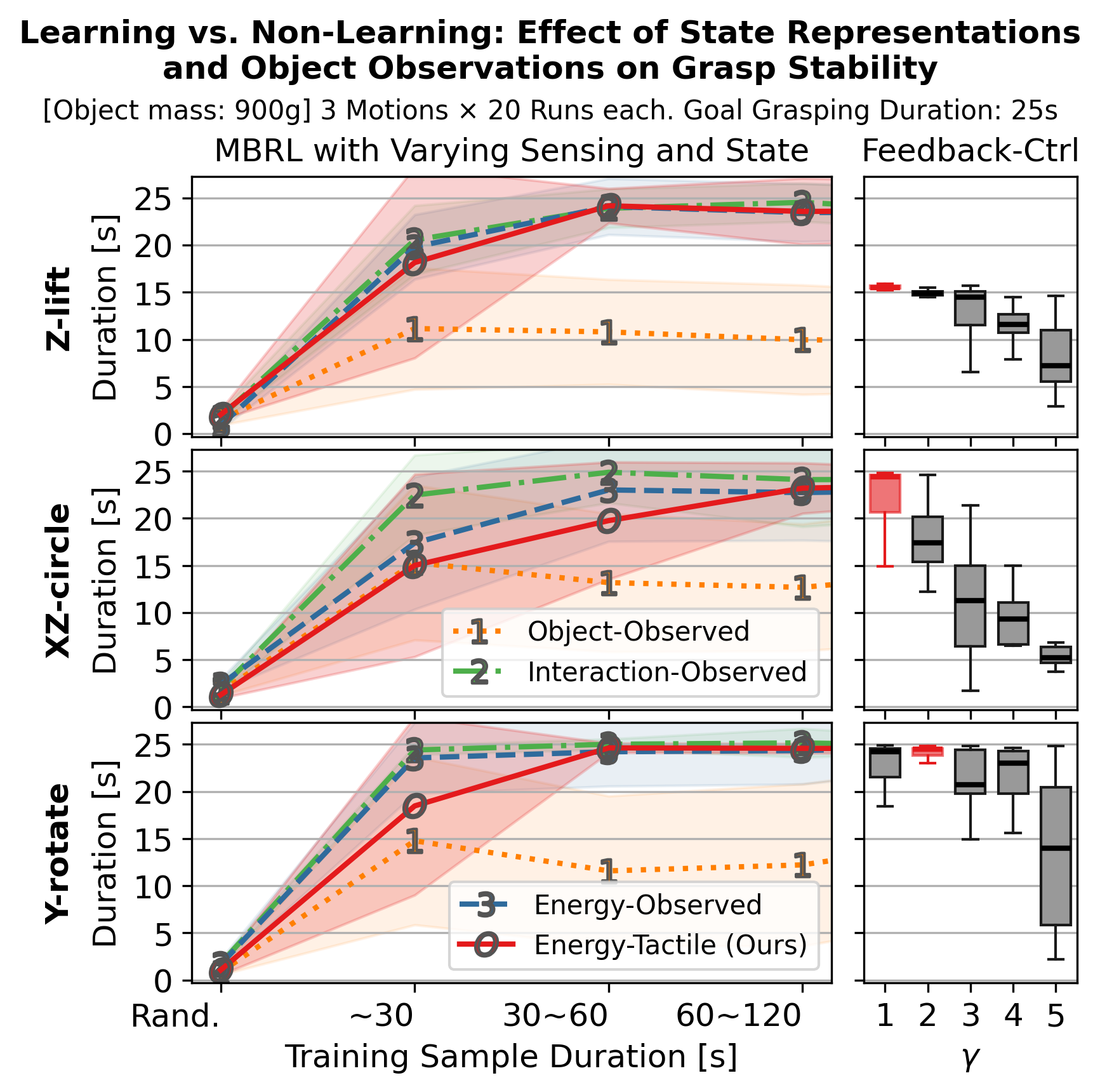}
    \caption{\textbf{Comparison of learning and non-learning baselines for grasping force control using a 900g object and three manipulation motions in simulation.} Learning-based methods differ in sensing and state representations, while ``Feedback-Ctrl'' applies a non-learning feedback strategy based on energy estimates (see Eq.~(\ref{eq:feedback_ctrol_force})).  Despite lacking object observation, ``Energy-Tactile (Ours)'' achieves comparable final performance, demonstrating the effectiveness of tactile-driven energy abstraction.}
    \label{fig:sim_baselines_comparison}
\end{figure}
\begin{figure*}
\centering
\includegraphics[width=2\columnwidth]{fig/sim_compare.png}
\caption{\textbf{Learning grasping force control for objects with unknown properties using only tactile sensing in simulation.} Experiments cover three object masses (100g, 500g, 900g) and three manipulation motions (Z-lift, XZ-circle, Y-rotate), each repeated 20 MBRL learn-from-scratch runs. Each row corresponds to a different motion type. \emph{Left}: Grasp duration increases as the learned dynamics improve, enabling more stable planning. \emph{Middle}: Mass estimates derived from the learned model converge toward ground truth, reflecting both model accuracy and the expressiveness of the energy abstraction. \emph{Right}: Grasping force consistently converges to distinct ranges for different object-motion pairs, highlighting the system’s ability to handle varying manipulation conditions. These results confirm that our MBRL framework can learn robust grasping force control in real time, without prior object knowledge or direct object observation.}
\label{fig:sim_compare}
\end{figure*}
\section{Setups for Experiment and Implementation}
\label{sec:setup_for_experiment}
\subsection{Experiment Environment}
We evaluate our method in both simulation and hardware using different dexterous robotic hands, demonstrating that it does not depend on a particular hand design (see Fig.~\ref{fig:robot_setup}).

\textbf{Hardware:} We use a Honda R\&D robotic hand~\cite{hasegawa2022icra}, which mimics an adult human hand and features four degrees of freedom per finger. Contact forces and positions are estimated using load cells at each joint, without dedicated fingertip tactile sensors. The robot is velocity controlled at 250Hz locally and communicates with MBRL at 10Hz.

Manipulated objects include two weighted 3D-printed shapes with smooth surfaces: a 320g ``Trapezoid'' with sloped edges that challenge stability, and a 350g ``Hammer'' with an offset center of mass, introducing dynamic balance challenges.

\textbf{Simulation:} We use a Shadow Dexterous Hand in MuJoCo~\cite{menagerie2022mujoco}, which is also velocity-controlled at 500Hz locally and communicates with MBRL at 10Hz in real-time.

The manipulated object is a 60mm$\times$40mm$\times$50mm cuboid with weights selected from \{0.1, 0.5, 0.9\} kg, aiming to isolate the impact of mass and inertia during manipulation. The robot must regulate grasping force to handle these properties.

\subsection{Key Evaluation Questions}
To help interpret the significance of our results, we frame our evaluation around the following questions:
\begin{enumerate}[leftmargin=*, label={\emph{Q\arabic*}}]
\item Why must the grasping force strategy account for different objects and manipulation motions?
\item Can energy abstraction serve as a reliable slip-aware control signal without external sensing?
% \item How effectively can MBRL with energy abstraction learn and handle different objects and manipulation motions using tactile and proprioceptive sensing only?
\item How effectively can MBRL with energy abstraction learn and handle different objects and manipulation motions?
\item Can our approach learn on hardware from scratch?
\end{enumerate}
The first three are examined through simulations, while the final question is explored through hardware experiments.

\subsection{Control Framework}
\label{sec:control_framework}
Our framework comprises three modules: nominal motion generation, grasping force optimization, and model trainer (see Fig.~\ref{fig:control_framework}). The nominal motion generator produces baseline Grasp Centroid (GC) and fingertip trajectories from a predefined motion library, inducing object motions such as grasping, lifting, and manipulation. Based on the learned energy dynamics, the grasping force optimizer controls fingertip forces in real time using pMPC. Motion and force commands are integrated and executed by a velocity-based impedance controller~\cite{hasegawa2022icra}. After each trial, the model trainer updates the dynamics model using the collected data. 

Each MBRL run consists of multiple trials, where each trial continues until the target grasp duration is reached (success) or the object is dropped (failure). In simulation, trials repeat until over two minutes of data (1200+ samples) are collected per run, typically yielding $\approx$15 trials. In hardware, each run includes a fixed set of eight trials due to scheduling constraints.

As some trials may terminate early, trial lengths can vary. Thus, all performance plots are aligned by accumulated sample duration rather than trial count.

\subsection{Nominal Manipulation Motion and Motion Phases}
Each MBRL trial comprises three phases:
\begin{enumerate}[leftmargin=*, label={[P\arabic*]}]
    \item \textbf{Approach}: a universal motion to approach the object: The thumb, index, and middle fingers move towards the GC until a contact force threshold is reached (2N for hardware, 5N for simulation; heuristically tuned to avoid false failure detection). Then, timer is reset ($t_a =0$) and \emph{pMPC grasping force optimization started}.
    \item \textbf{Lift}: Elevate the object by $l_{init}$\,mm.
    \item \textbf{Manipulation}: Execute one periodic motion from:
\end{enumerate}
\begin{enumerate}[leftmargin=47pt,label={[M\arabic*]}]
    \item \emph{Z-lift}: Vertical sinusoidal lifting.
    \item \emph{XZ-circle}: Circular motion in the xz-plane.
    \item \emph{Y-rotate}: Periodic pitch rotation about the y-axis.
\end{enumerate}
Motion details are provided in Fig.~\ref{fig:predefined_motion}. Phase changes at $t_a = 0.5$s (P1$\rightarrow$P2) and $t_a = 4$s (P2$\rightarrow$P3). Phase P3 terminates at $t_a = 25$s for simulation and $t_a = 20$s for hardware.

\subsection{Real-time Grasping Force Planning via pMPC}
We use pMPC to minimize fluctuations in estimated mass by defining the immediate loss function as:
\begin{align}
    \label{eq:immediate_loss}
    \func{\ell}{}{\bst{x}{k}} = -\exp \sbrkt{-\alpha|\tilde{m}_k - \mathrm{med}\sbrkt{\tilde{m}_{t}}|},
\end{align}
where $\mathrm{med}(\tilde{m}_{t})$ is the median of estimated masses up to time $t$, and $\alpha = 200$.
To avoid vibration, the pMPC control space is constrained to a $\pm3$N window on the previously applied force: $F_{G,t} \in \mbrkt{F_{G,t-1} \pm 3\text{N}}$. Force limits are set to $[0, 30]$N in simulation and $[0, 20]$N in hardware. During the first MBRL trial, a random grasp force is sampled from $[0, 20]$N (simulation) or $[0, 10]$N (hardware).

%% file: content/6_exp_result.tex
\section{Experimental Results}\label{sec:experiment}
\qsubsection{Why must the grasping force strategy account for different objects and manipulation motions?}
To investigate how grasping force affects slippage under varying conditions, we conduct a pre-learning analysis using fixed-force trials. As shown in Fig.~\ref{fig:sim_slippage_force_compare}, each object-motion pair exhibits a distinct relationship between grasping force and slippage, and the optimal force for minimizing slippage varies significantly with object mass and motion type. For lighter objects (left), excessive force leads to instability, while heavier objects (right) demand substantially more force to avoid failure. These results show that the optimal grasping force varies with each object-motion pair and cannot be determined without case-specific trial-and-error. This underscores the need for adaptive force strategies, as no single predefined value can handle diverse interaction conditions reliably.

\qsubsection{Can energy abstraction serve as a reliable slip-aware control signal without external sensing?}
To assess whether the energy abstraction reflects slippage, we analyze energy-based mass estimates under fixed-force trials. As shown in Fig.~\ref{fig:sim_slippage_mass_compare}, successful trials yield accurate, consistent estimates near the true mass, while failing trials show greater slippage and wider deviation. These results support energy abstraction as a physically grounded signal for grasp stability without external sensing.

Next, we empirically evaluate five configurations using the same object (900g) across all three manipulation motions:
\begin{enumerate}[leftmargin=*]
    \item \textbf{Feedback-Ctrl}: A non-learning baseline that adjusts grasping force based on energy-based mass estimates as:
    \begin{equation}
        F_{G,t} = F_{G,t-1} + \gamma(\tilde{m}_t - \text{med}(\tilde{m}_{t}))\cdot 3\text{N},
        \label{eq:feedback_ctrol_force}
    \end{equation}
    where we tested the gain term across $\gamma \in \lbrkt{1, 2, 3, 4, 5}$. This verifies whether stable grasping can be achieved using simple feedback without learning or prediction.
    \item \textbf{Object-Observed} (MBRL with external object state sensing): The state includes only the object’s motion (no finger interaction), and pMPC simply maximizes the object’s z-height. This naive baseline tests whether object-level heuristics suffice for maintaining stability.
    \item \textbf{Interaction-Observed} (MBRL with external object state observing): The state includes both the object’s motion and grasp centroid (GC), and pMPC minimizes the discrepancy between them to reduce slippage. This serves as an upper-bound baseline under full observability.
    \item \textbf{Energy-Observed} (MBRL with external object state observing): The state and pMPC objective are identical to our approach, but energy features are computed directly from object observation, serving as a reference to validate our energy abstraction in low-uncertainty settings.
    \item \textbf{Energy-Tactile (Ours)}: Our proposed approach uses only tactile sensing to learn energy-based interaction dynamics and optimize grasping force.
\end{enumerate}

All results are shown in Fig.~\ref{fig:sim_baselines_comparison}. \textbf{Feedback-Ctrl} demonstrates that energy-based mass estimates can encode slip-relevant information, enabling stabilization for specific motions and gain values. However, the optimal $\gamma$ varies by condition, and in \emph{Z-lift}, manual tuning fails to achieve the target 25s grasp duration, highlighting the limited generalizability of this approach and the need for learning-based strategies. \textbf{Object-Observed} also fails to stabilize the object under dynamic motions, revealing the limitations of heuristic-based learning. In contrast, other learning-based methods, adapt robustly across object-motion variations. While \textbf{Interaction-Observed} performs best under full observability, \textbf{Energy-Observed} achieves comparable results, validating the energy abstraction itself. Most notably, \textbf{Energy-Tactile} achieves similar final performance using only tactile sensing, with minimal learning delay, demonstrating the effectiveness and practicality of our method under sensing constraints.

\qsubsection{How effectively can MBRL with energy abstraction learn and handle different objects and manipulation motions?}
We conducted learn-from-scratch experiments across three object masses, three manipulation motions, and 20 MBRL runs per object-motion pair. As shown in Fig.\ref{fig:sim_compare}-Left, our method rapidly improves grasp duration within two minutes of interaction. Without external sensing or prior knowledge of object mass, the learned model accurately captures object dynamics, evidenced by the fast convergence of energy-based mass estimates toward ground truth (Fig.\ref{fig:sim_compare}-Middle). Finally, the grasping force distributions consistently converge to distinct ranges for each object-motion pair (Fig.~\ref{fig:sim_compare}-Right), demonstrating that our approach generalizes robustly across diverse conditions using only tactile sensing.

In addition, we present an example trial in Fig.~\ref{fig:sim_learning_example}. The object trajectory closely follows the desired path, resulting in a stable grasp with less than 8mm of cumulative slippage over 25s. Notably, the grasp force exhibits a quasi-periodic pattern reflecting the intended \emph{Z-lift} motion, varying between 5N and 25N for this object-motion setting. This variation in force is also reflected in the summarize results across all conditions (see Fig.~\ref{fig:sim_compare}-Right). This example demonstrates how our method uses energy consistency to adaptively optimize grasping force.

\qsubsection{Can our approach learn on hardware from scratch?}
We evaluate our method on hardware using two objects with unknown properties (``Trapezoid'' and ``Hammer'') across three manipulation motions. As shown in Fig.~\ref{fig:real_compare}, our method consistently improves grasp duration within one minute of interaction, despite learning-from-scratch using only tactile sensing. The estimated object masses converge toward ground truth values, validating the applicability of learning energy dynamics in real-world. Similar to simulated results, the learned grasping forces adaptively converge to higher values for the heavier and geometrically more challenging ``Hammer,'' demonstrating our method’s ability to learn and regulate grasping force directly on hardware.

%% file: content/7_discussion.tex
\begin{figure}[!t]
    \centering
    \includegraphics[width=1\linewidth]{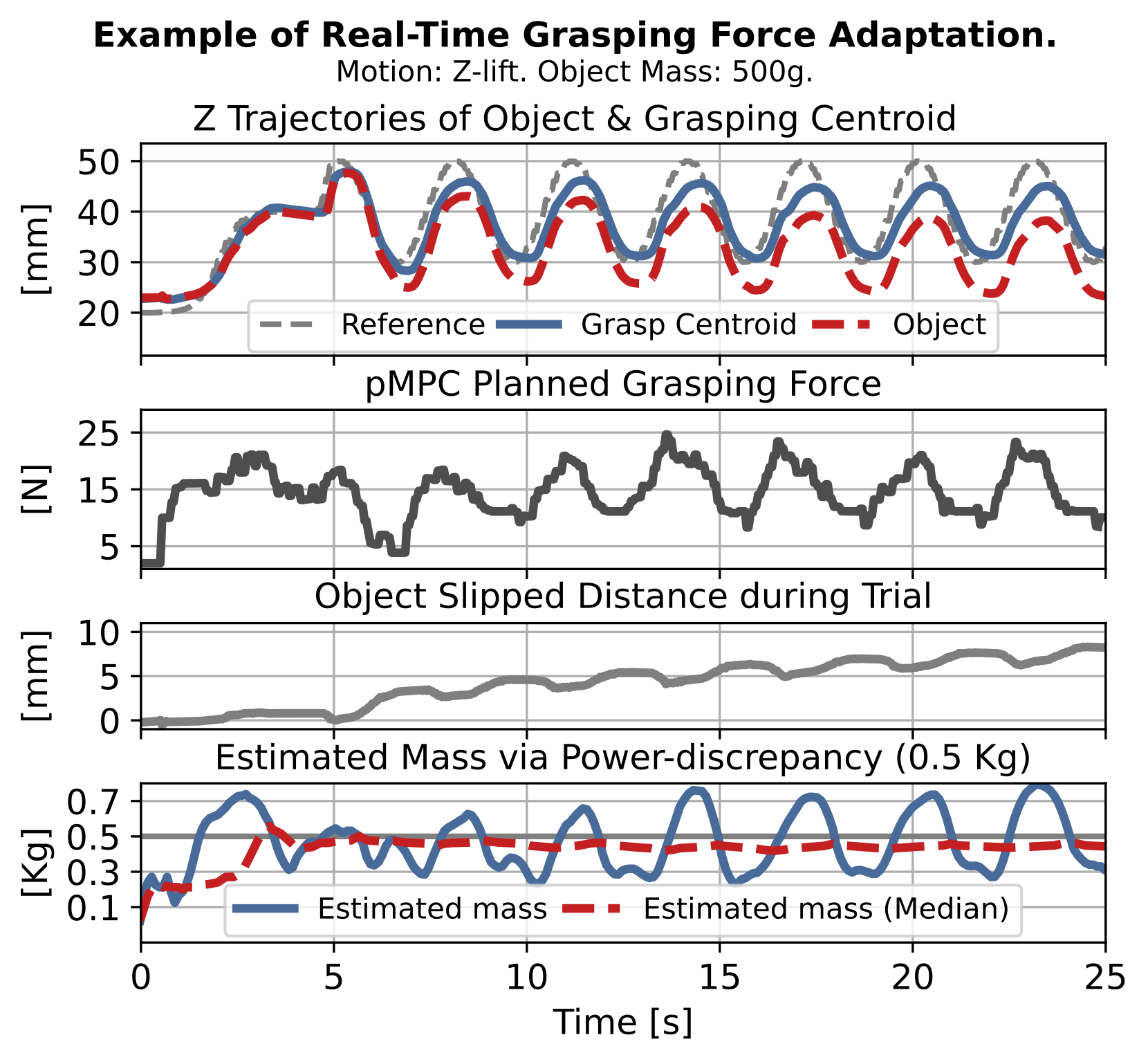}
    \caption{\textbf{An example trial showing real-time grasping force control under the proposed MBRL framework in simulation (motion: Z-lift; object: 500g).} 
    \emph{Top}: Object and grasp centroid trajectories remain closely aligned throughout motion. \emph{Second}: Grasping force is continuously adapted in real time to accommodate changes in interaction dynamics induced by the periodic ``Z-Lift'' motion. \emph{Third}: Cumulative slippage stays below 8 mm during the trial. \emph{Bottom}: Energy-based mass estimation remains stable, indicating consistent learning of object interaction dynamics.}
    
    \label{fig:sim_learning_example}
\end{figure}

\begin{figure*}
\centering
\includegraphics[width=2\columnwidth]{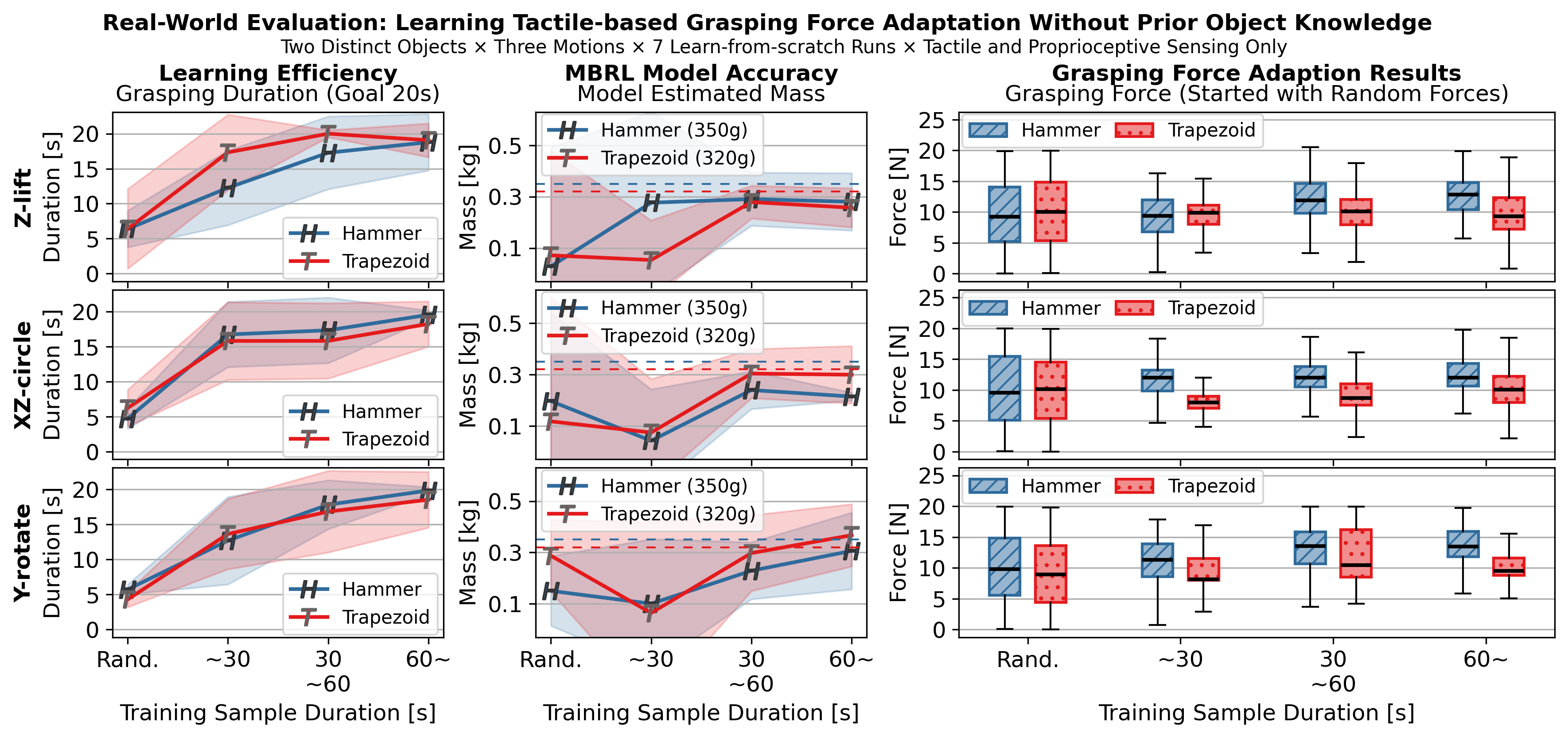}
\caption{\textbf{Real-world evaluation of tactile-based grasping force control without prior object knowledge.} The robot learns grasping force control on two unseen object across three motions. \emph{Left}: Grasp duration increases within one minute. \emph{Middle}: Mass estimates approach ground truth (dashed lines), reflecting accurate energy dynamics learning. \emph{Right}: Grasp forces converge to motion-specific values. The full learning process (trials, resets, and training) averages $512\pm30.4$s. These results demonstrate the system’s ability to learn tactile-based force control on hardware for slippage reduction.}
\label{fig:real_compare}
\end{figure*}

\section{Discussion}\label{sec:discussion}
While our method demonstrates effective tactile-based grasping force control, it has limitations and requirements. Our method is restricted to rigid objects with fixed mass. It also requires 3D fingertip contact forces and positions, which can be obtained directly by force sensors or joint load cells (as in this work). With recent advances in tactile signal processing, we believe our method can be extended to robots that use tactile sensors to estimate the required contact information.

Future work may extend our modular framework to advanced manipulation tasks (such as contact shifts, palm engagement, or intricate motions) by integrating RL policies to generate reference trajectories or to replace pMPC planning. While pMPC provides high stability, its computational cost can limit control frequency and scalability. Combining MBRL with RL policies, and leveraging energy abstraction as a reward or feedback, may further improve efficiency, interpretability, and capability for advanced manipulation.

%% file: content/8_conclusion.tex
\section{Conclusion}\label{sec:conclusion}
We presented a physics-informed energy abstraction that enables tactile-based grasping force control to reduce slippage during dynamic object interaction. By leveraging energy consistency, the approach provides a compact and physically grounded signal for slip-aware reasoning without requiring object priors or explicit supervision. Integrated with model-based reinforcement learning, the system achieves rapid learning and real-time control in both simulation and hardware across diverse object-motion pairs, demonstrating its potential for sample-efficient tactile manipulation.

%% file: src/library.bib
@inproceedings{kamthe2018aistats,
  author    = {Sanket Kamthe and
               Marc Peter Deisenroth},
  title     = {Data-Efficient Reinforcement Learning with Probabilistic Model Predictive
               Control},
  booktitle = {Int. Conf. Artificial Intelligence and Statistics},
  volume    = {84},
  pages     = {1701-1710},
  year      = {2018},
}

@inproceedings{deisenroth2011icml,
 author     = {Deisenroth, Marc Peter and Rasmussen, Carl Edward},
 title      = {PILCO: A Model-based and Data-efficient Approach to Policy Search},
 booktitle  = {Int. Conf. Machine Learning},
 year       = {2011},
 pages      = {465-472},
}

@article{kuo2023ral,
  author={Kuo, Cheng-Yu and Shin, Hirofumi and others},
  journal={IEEE Robot. Automat. Lett.}, 
  title={Reinforcement Learning With Energy-Exchange Dynamics for Spring-Loaded Biped Robot Walking}, 
  year={2023},
  volume={8},
  number={10},
  pages={6243-6250},
}

@misc{menagerie2022mujoco,
  author = {Zakka, Kevin and Tassa, Yuval and others},
  title = {{MuJoCo Menagerie: A collection of high-quality simulation models for MuJoCo}},
  url = {http://github.com/google-deepmind/mujoco_menagerie},
  year = {2022},
}

@inproceedings{hasegawa2022icra,
  author={Hasegawa, Tadaaki and Waita, Hironori and others},
  booktitle={IEEE Int. Conf. Robot. Autom.}, 
  title={Powerful and dexterous multi-finger hand using dynamical pulley mechanism}, 
  year={2022},
  volume={},
  number={},
  pages={707-713},
}

@article{roa2014autonomous,
  author = {Roa, M. and Suárez, R.},
  title = {grasp quality measures: review and performance},
  journal = {Autonomous Robots},
  year = {2014},
  volume = {38},
  issue = {1},
  pages = {65-88},
  doi = {10.1007/s10514-014-9402-3}
}

@inproceedings{ferrari1992icra,
  author={Ferrari, C. and others},
  booktitle={IEEE Int. Conf. Robot. Autom.}, 
  title={Planning optimal grasps}, 
  year={1992},
  volume={},
  number={},
  pages={2290-2295 vol.3},
  doi={10.1109/ROBOT.1992.219918}}

@article{melnik2021froroboticsandai,
  author = {Melnik, A. and Lach, L. and others},
  title = {Using tactile sensing to improve the sample efficiency and performance of deep deterministic policy gradients for simulated in-hand manipulation tasks},
  journal = {Front. Robot. AI},
  year = {2021},
  volume = {8},
  doi = {10.3389/frobt.2021.538773}
}

@article{andrychowicz2019ijrr,
  author = {Andrychowicz, O. M. and Baker, B. and others},
  title = {Learning dexterous in-hand manipulation},
  journal = {Int. J. Robot. Res.},
  year = {2019},
  volume = {39},
  issue = {1},
  pages = {3-20},
  doi = {10.1177/0278364919887447}
}

@inproceedings{abd2018aim,
  author={Abd, Moaed A. and Gonzalez, Iker J. and others},
  booktitle={IEEE/ASME Int. Conf. Advanced Intelligent Mechatronics}, 
  title={Direction of Slip Detection for Adaptive Grasp Force Control with a Dexterous Robotic Hand}, 
  year={2018},
  volume={},
  number={},
  pages={21-27},
  doi={10.1109/AIM.2018.8452704}}

@INPROCEEDINGS{bierbaum2008humanoids,
  author={Bierbaum, Alexander and Rambow, Matthias and Asfour, Tamim and Dillmann, Rudiger},
  booktitle={IEEE-RAS Int. Conf. Humanoid Robots}, 
  title={A potential field approach to dexterous tactile exploration of unknown objects}, 
  year={2008},
  volume={},
  number={},
  pages={360-366},
  doi={10.1109/ICHR.2008.4756005}}

@INPROCEEDINGS{song2018robio,
  author={Song, Fangjing and Zhao, Zengzhi and others},
  booktitle={IEEE Int. Conf. Robotics and Biomimetics}, 
  title={Learning Optimal Grasping Posture of Multi-Fingered Dexterous Hands for Unknown Objects}, 
  year={2018},
  volume={},
  number={},
  pages={2310-2315},
  doi={10.1109/ROBIO.2018.8665277}}

@INPROCEEDINGS{morgan2019iros,
  author={Morgan, Andrew S. and Hang, Kaiyu and others},
  booktitle={IEEE/RSJ Int. Conf. Intell. Robots Syst.}, 
  title={A Data-Driven Framework for Learning Dexterous Manipulation of Unknown Objects}, 
  year={2019},
  volume={},
  number={},
  pages={8273-8280},
  doi={10.1109/IROS40897.2019.8968458}}

@INPROCEEDINGS{veiga2015iros,
  author={Veiga, Filipe and others},
  booktitle={IEEE/RSJ Int. Conf. Intell. Robots Syst.}, 
  title={Stabilizing novel objects by learning to predict tactile slip}, 
  year={2015},
  volume={},
  number={},
  pages={5065-5072},
  doi={10.1109/IROS.2015.7354090}}

@INPROCEEDINGS{su2015humanoids,
  author={Su, Zhe and Hausman, Karol and others},
  booktitle={IEEE-RAS Int. Conf. Humanoid Robots}, 
  title={Force estimation and slip detection/classification for grip control using a biomimetic tactile sensor}, 
  year={2015},
  volume={},
  number={},
  pages={297-303},
  doi={10.1109/HUMANOIDS.2015.7363558}}

@ARTICLE{chen2024dexforce,
  author={Chen, Claire and Yu, Zhongchun and others},
  journal={IEEE Robot. Autom. Lett.}, 
  title={DexForce: Extracting Force-Informed Actions From Kinesthetic Demonstrations for Dexterous Manipulation}, 
  year={2025},
  volume={10},
  number={6},
  pages={6416-6423},
}

@article{openai2019rubiks,
  title={Solving Rubik's Cube with a Robot Hand},
  author={OpenAI and Andrychowicz, Marcin and others},
  journal={arXiv preprint arXiv:1910.07113},
  year={2019}
}

@INPROCEEDINGS{funabashi2025focused,
  author={Funabashi, Satoshi and others},
  booktitle={IEEE Int. Conf.  Robot. Autom.}, 
  title={Focused Blind Switching Manipulation Based on Constrained and Regional Touch States of Multi-Fingered Hand Using Deep Learning}, 
  year={2025},
  volume={},
  number={},
  pages={14777-14783},
}

@INPROCEEDINGS{jiang2024contactmpc,
  author={Jiang, Yongpeng and Yu, Mingrui and others},
  booktitle={IEEE/RSJ Int. Conf. Intell. Robots Syst.}, 
  title={Contact-Implicit Model Predictive Control for Dexterous In-hand Manipulation: A Long-Horizon and Robust Approach}, 
  year={2024},
  volume={},
  number={},
  pages={5260-5266},
}

@ARTICLE{morgan2022ral,
  author={Morgan, Andrew S. and Hang, Kaiyu and others},
  journal={IEEE Robot. Autom. Lett.}, 
  title={Complex In-Hand Manipulation Via Compliance-Enabled Finger Gaiting and Multi-Modal Planning}, 
  year={2022},
  volume={7},
  number={2},
  pages={4821-4828}}

@ARTICLE{komeno2024ral,
  author={Komeno, Naoto and Matsubara, Takamitsu},
  journal={IEEE Robot. Autom. Lett.}, 
  title={Incipient Slip Detection by Vibration Injection into Soft Sensor}, 
  year={2024},
  volume={9},
  number={4},
  pages={3251-3258}
}

@INPROCEEDINGS{sundaralingam2018icra,
  author={Sundaralingam, Balakumar and Hermans, Tucker},
  booktitle={IEEE Int. Conf.  Robot. Autom.}, 
  title={Geometric In-Hand Regrasp Planning: Alternating Optimization of Finger Gaits and In-Grasp Manipulation}, 
  year={2018},
  volume={},
  number={},
  pages={231-238},
}

@INPROCEEDINGS{valencia2023icra,
  author={Valencia, David and Jia, John and others},
  booktitle={IEEE Int. Conf. Robot. Autom.}, 
  title={Comparison of Model-Based and Model-Free Reinforcement Learning for Real-World Dexterous Robotic Manipulation Tasks}, 
  year={2023},
  volume={},
  number={},
  pages={871-878},
  doi={10.1109/ICRA48891.2023.10160983}}

@INPROCEEDINGS{kumar2016icra,
  author={Kumar, Vikash and Todorov, Emanuel and Levine, Sergey},
  booktitle={IEEE Int. Conf. Robot. Autom.}, 
  title={Optimal control with learned local models: Application to dexterous manipulation}, 
  year={2016},
  volume={},
  number={},
  pages={378-383}}

@INPROCEEDINGS{nagabandi2019corl, 
     AUTHOR = {Nagabandi, Anusha and  Konoglie, Kurt and others}, 
     TITLE = "{Deep Dynamics Models for Learning  Dexterous Manipulation}", 
     BOOKTITLE = {Conference on Robot Learning}, 
     YEAR = {2019}, }

@INPROCEEDINGS{omer2021iceee,
  author={Omer, Muhammad and Ahmed, Rami and others},
  booktitle={Int. Conf.  Computer, Control, Electrical, and Electronics Engineering}, 
  title={Model Predictive-Actor Critic Reinforcement Learning for Dexterous Manipulation}, 
  year={2021},
  volume={},
  number={},
  pages={1-6}}

@INPROCEEDINGS{radosavovic2021iros,
  author={Radosavovic, Ilija and Wang, Xiaolong and others},
  booktitle={IEEE/RSJ Int. Conf. Intell. Robots Syst.}, 
  title={State-Only Imitation Learning for Dexterous Manipulation}, 
  year={2021},
  volume={},
  number={},
  pages={7865-7871},
  doi={10.1109/IROS51168.2021.9636557}}

@article{schwarz2016tn,
title = {The Slip Hypothesis: Tactile Perception and its Neuronal Bases},
journal = {Trends in Neurosciences},
volume = {39},
number = {7},
pages = {449-462},
year = {2016},
author = {Cornelius Schwarz},
}

@INPROCEEDINGS{karl2018icra,
  author={Van Wyk, Karl and Falco, Joe},
  booktitle={IEEE Int. Conf. Robot. Autom.}, 
  title={Calibration and Analysis of Tactile Sensors as Slip Detectors}, 
  year={2018},
  volume={},
  number={},
  pages={2744-2751},
  doi={10.1109/ICRA.2018.8461117}}

@article{folkertsma2017jtr,
title = "Energy in Robotics",
author = "Folkertsma, {Gerrit A.} and Stefano Stramigioli",
year = "2017",
volume = "6",
pages = "140-210",
journal = "Foundations and Trends{\textregistered} in Robotics",
publisher = "Now Publishers Inc",
number = "3",
}

@ARTICLE{damian2015asme,
  author={Damian, Dana D. and Newton, Taylor H. and others},
  journal={IEEE/ASME Trans. Mechatron.}, 
  title={Artificial Tactile Sensing of Position and Slip Speed by Exploiting Geometrical Features}, 
  year={2015},
  volume={20},
  number={1},
  pages={263-274},
}

@article{trinkle1990ijrr,
author = {J.C. Trinkle and R.P. Paul},
title ={Planning for Dexterous Manipulation with Sliding Contacts},
journal = {Int. J. Robot. Res.},
volume = {9},
number = {3},
pages = {24-48},
year = {1990},
doi = {10.1177/027836499000900302},
}
